\documentclass[10pt]{article} 
\usepackage[preprint]{tmlr}

\usepackage{amsmath,amsfonts,bm}

\def\eqref#1{equation~\ref{#1}}

\def\1{\bm{1}}

\DeclareMathAlphabet{\mathsfit}{\encodingdefault}{\sfdefault}{m}{sl}
\SetMathAlphabet{\mathsfit}{bold}{\encodingdefault}{\sfdefault}{bx}{n}

\DeclareMathOperator*{\argmax}{arg\,max}

\usepackage{hyperref}
\usepackage{url}
\usepackage{xcolor}
\usepackage{graphicx} 
\usepackage{array} 
\usepackage{caption}  
\usepackage{booktabs} 
\usepackage[ruled,lined]{algorithm2e}
\usepackage{algorithmic}
\definecolor{lightblue}{RGB}{57,113,181}
\usepackage{multicol}
\usepackage{graphicx} 
\usepackage{wrapfig} 
\usepackage{lipsum} 
\usepackage{paralist}
\newcommand{\fedmmeMethod}{\texttt{FedMME}}
\newcommand{\myhline}
{\noalign{\global\arrayrulewidth0.3mm}\hline
                      \noalign{\global\arrayrulewidth0.3pt}}

\title{Multi-Modal One-Shot Federated Ensemble Learning for Medical Data with Vision Large Language Model}

\author{\name Naibo Wang \email naibowang@u.nus.edu \\
      \addr Institute of Data Science\\
      National University of Singapore
      \AND
      \name Yuchen Deng \email dengyuchen.cc@gmail.com \\
      \addr School of Mathematics and Statistics\\
      Changchun University of Technology
      \AND
      \name Shichen Fan \email shichenfan@stu.xidian.edu.cn\\
      \addr School of Computer Science and Technology \\
      Xidian University
      \AND
      \name Jianwei Yin \email zjuyjw@cs.zju.edu.cn\\
      \addr College of Computer Science and Technology \\
      Zhejiang University
      \AND
      \name See-Kiong Ng \email seekiong@nus.edu.sg\\
      \addr Institute of Data Science \\
      National University of Singapore
      }

\def\month{MM}  
\def\year{YYYY} 
\def\openreview{\url{https://openreview.net/forum?id=XXXX}} 

\begin{document}

\maketitle

\begin{abstract}
Federated learning (FL) has attracted considerable interest in the medical domain due to its capacity to facilitate collaborative model training while maintaining data privacy. However, conventional FL methods typically necessitate multiple communication rounds, leading to significant communication overhead and delays, especially in environments with limited bandwidth. One-shot federated learning addresses these issues by conducting model training and aggregation in a single communication round, thereby reducing communication costs while preserving privacy. Among these, one-shot federated ensemble learning combines independently trained client models using ensemble techniques such as voting, further boosting performance in non-IID data scenarios. On the other hand, existing machine learning methods in healthcare predominantly use unimodal data (e.g., medical images or textual reports), which restricts their diagnostic accuracy and comprehensiveness. Therefore, the integration of multi-modal data is proposed to address these shortcomings. Additionally, vision large language models (vLLMs) have emerged as powerful tools due to their ability to interpret and generate textual descriptions from visual data, making them invaluable for creating textual reports from medical images. In this paper, we introduce \fedmmeMethod, an innovative one-shot multi-modal federated ensemble learning framework that utilizes multi-modal data for medical image analysis.  Specifically, \fedmmeMethod\ capitalizes on vision large language models to produce textual reports from medical images, employs a BERT model to extract textual features from these reports, and amalgamates these features with visual features to improve diagnostic accuracy. Experimental results show that our method demonstrated superior performance compared to existing one-shot federated learning methods in healthcare scenarios across four datasets with various data distributions. For instance, it surpasses existing one-shot federated learning approaches by more than 17.5\% in accuracy on the RSNA dataset when applying a Dirichlet distribution with ($\alpha$ = 0.3).
\end{abstract}

\section{Introduction}

\begin{figure*}[t]
    \centering
     \includegraphics[width=\linewidth]{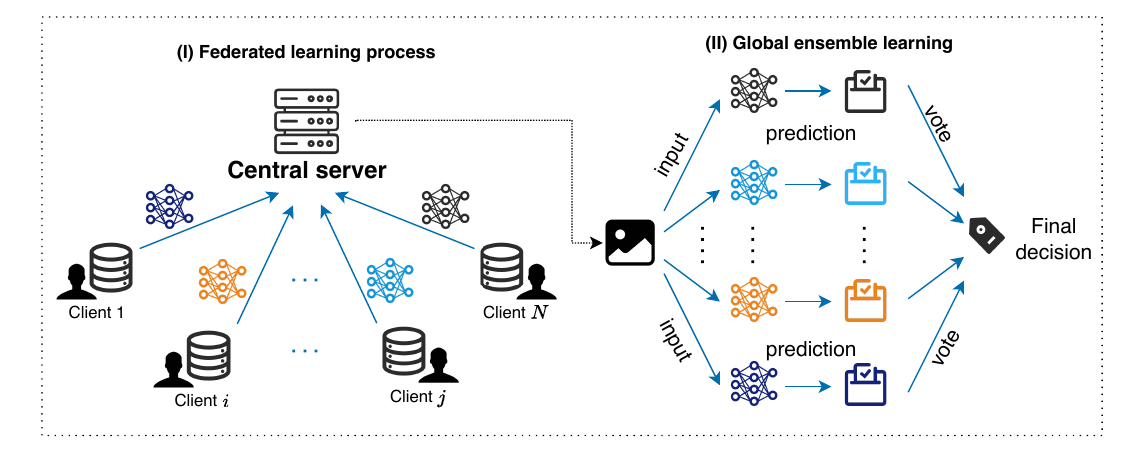}
     \caption{Overview of the one-shot federated ensemble framework. (I) Federated learning process: each client trains a local model using its private dataset and transmits the model to the central server. (II) Global ensemble learning: the server aggregates model outputs from all clients using a voting mechanism to produce the final decision.}

     \label{fig:overview}
\end{figure*}

Federated learning~\citep{mcmahan2017communication} is a paradigm that allows multiple distributed clients to collaboratively train a global model without sharing their local data~\citep{wang2022collaborative, balkus2022survey}. This approach encompasses various scenarios such as parallel federated learning~\citep{li2020federated}, sequential federated learning~\citep{wang2024one}, and federated ensemble learning~\citep{wang2023data}. As depicted in Fig. \ref{fig:overview}, federated ensemble learning involves each participant independently training a model on their respective local datasets. Subsequently, these models are combined on a central server to perform ensemble learning techniques, including voting~\citep{raza2019improving} or stacking~\citep{cui2021stacking}. Federated ensemble learning has gained substantial popularity in privacy-sensitive areas including finance~\citep{gadekallu2021federated} and edge computing~\citep{alam2023federated}.

Recently, federated learning has gained significant attention in the medical field~\citep{pfitzner2021federated, sheller2020federated} because it enables the development of more comprehensive and accurate models by integrating data from different medical institutions without compromising data privacy. In contrast, traditional centralized machine learning~\citep{drainakis2020federated} requires data aggregation on a central server for training, which presents substantial privacy, legal, and security challenges when handling sensitive medical data. Federated learning mitigates these challenges by allowing multiple medical institutions to perform model training in their local environments, exchanging only model updates rather than raw data. This method effectively safeguards data privacy and produces a robust global model or ensemble team that surpasses any model developed by individual institutions alone~\citep{abaoud2023advancing}.

Despite the effectiveness of federated learning in protecting privacy and enabling distributed training for healthcare, it faces challenges related to substantial communication overhead and time delays, especially in environments with limited network bandwidth~\citep{kim2024fedwt}. To address the risks of attack and high communication costs associated with traditional federated learning, the concept of one-shot federated learning has been proposed~\citep{guha2018one, dennis2021heterogeneity}. This communication-efficient method conducts model training and aggregation in a single communication round, significantly reducing overhead and enhancing data privacy~\citep{zhang2022dense}. Consequently, employing one-shot federated ensemble learning for the development of healthcare applications holds considerable promise. 

In the medical field, federated learning methods commonly rely on unimodal data for training, such as solely using medical images~\citep{chakravarty2021federated, ke2021style, kang2023one} or textual reports~\citep{sui2020feded}. However, this unimodal approach results in a significant loss of valuable information. For example, analyzing medical images without the context provided by patient history or textual reports can lead to incomplete diagnostics or overlook critical correlations between visual cues and underlying conditions~\citep{reale2024vision}. This information bottleneck restricts a comprehensive understanding and accurate interpretation of a patient's health status. 


Consider a scenario in which a CT scan detects an anomaly that initially appears benign. Without incorporating textual data, such as the patient's symptoms or genetic predisposition from their medical history, physicians may overlook the need for further investigation of a potentially serious condition~\citep{rakowski2006renal}. Similarly, when relying solely on textual reports, subtle indicators in the imagery might be entirely missed~\citep{wu2024xlip}. By integrating various data types—such as medical images, textual reports, and sensory data from wearable devices—healthcare models can obtain a more comprehensive understanding of patient health. This integrative approach enables the cross-validation and enrichment of datasets~\citep{singh2020explainable}, leading to insights that cannot be achieved through the analysis of unimodal data alone. Therefore, incorporating multi-modal data into one-shot federated learning frameworks for healthcare applications is imperative to achieve more accurate and reliable diagnostic and predictive outcomes.


In recent years, large language models (LLMs) have demonstrated substantial potential in natural language processing~\citep{brown2020language,chowdhery2023palm,hoffmann2022training} and multi-modal~\citep{hu2024bliva,ge2024worldgpt} tasks. These models excel in generating high-quality textual reports and in extracting and integrating complex semantic information from both textual and visual data, particularly when augmented with their vision-based counterparts, such as Llama-3.2-11B-Vision~\citep{chi2024llama}. The advanced capabilities of LLMs offer new opportunities for multi-modal analysis in the medical domain~\citep{ghosh2024clipsyntel}. A medical analysis framework incorporating LLMs can produce detailed imaging reports, thereby enhancing diagnostic accuracy and interoperability~\citep{waldock2024accuracy}. Hence, utilizing these advanced vision large language models within medical settings could substantially improve therapeutic effects and streamline clinical workflows.

In this paper, we introduce \fedmmeMethod, an innovative \textit{Fed}erated \textit{M}ulti-\textit{M}odal \textit{E}nsemble learning framework for medical image analysis. Our framework enhances medical data analysis by incorporating both visual and textual features, with the help of vision large language models. Specifically, our method employs these models to generate textual descriptions from medical images, which are then integrated with visual data for model training. This approach significantly enhances prediction accuracy and robustness in medical image analysis. Additionally, our framework effectively harnesses multi-modal information while maintaining low communication costs, making it particularly suited for privacy-sensitive and resource-constrained medical applications such as medical image analysis.

Our main contributions can be summarized as follows:
\begin{compactitem}[$\bullet$]

    \item We introduce \fedmmeMethod, a novel one-shot, multi-modal federated learning framework designed for medical image analysis. To the best of our knowledge, this is the first work to systematically explore the one-shot federated ensemble learning for training multi-modal models.
    \item We innovatively employ large vision language models to generate more accurate medical reports from images, thereby extracting superior textual features that enhance overall performance.
    \item We conduct comprehensive experiments across four datasets with various data distributions. Our method demonstrates superior performance compared to existing one-shot FL methods in healthcare scenarios.
\end{compactitem}
\section{Related Work}

Federated Learning (FL)~\citep{mcmahan2017communication} is a decentralized machine learning framework that aims to train models without centralizing user data, thus preserving privacy and reducing communication costs. In recent years, FL has made significant advances in addressing data heterogeneity~\citep{li2020federated,karimireddy2020scaffold,ye2023feddisco}, enhancing privacy protection~\citep{geyer2017differentially,wei2020federated}, and optimizing communication efficiency~\citep{caldas2018expanding}. Despite these advancements, traditional FL often necessitates multiple rounds of communication to achieve global optimization, presenting challenges in communication-limited environments. To overcome this, one-shot federated learning has been introduced~\citep{guha2019one,su2023one}, which accomplishes model aggregation in a single round of global communication, thereby significantly reducing communication overhead. Specifically, FedDISC~\citep{yang2024exploring} investigates one-shot semi-supervised federated learning using a pre-trained diffusion model. Additionally, DENSE~\citep{zhang2022dense} presents a data-free one-shot federated learning approach through knowledge distillation. FedISCA~\citep{kang2023one} is the state-of-the-art one-shot federated learning framework for medical applications, which effectively addresses data heterogeneity by generating synthetic data and adapting client models with the help of knowledge distillation. All these methods support the rapid deployment of federated models in scenarios where frequent communication or data sharing is impractical or undesirable, often due to privacy considerations or bandwidth constraints.

Ensemble learning seeks to combine multiple weak base models to create a more robust model. This approach has been extensively researched for decades and is applicable in various contexts, including federated learning. Traditional ensemble learning techniques include Voting~\citep{raza2019improving}, Bagging~\citep{breiman1996bagging}, Boosting~\citep{schapire2013explaining}, and Stacking~\citep{wolpert1992stacked}. Federated Ensemble Learning~\citep{wang2023data} extends these traditional techniques to the decentralized setting of federated learning. FedDF~\citep{lin2020ensemble} introduces ensemble distillation, a method for robustly fusing heterogeneous client models in federated learning by training a central model on unlabeled data using client model predictions. FedEL~\citep{wu2024fedel} introduces a federated ensemble learning approach that trains diverse weak learners across non-IID client data and combines them into a robust global model to improve performance under data heterogeneity. In this study, we explore the training schemes of multi-modal models in federated ensemble learning.


Large Language Models (LLMs)~\citep{brown2020language, chowdhery2023palm, ouyang2022training} have garnered recognition for their advanced capabilities in Natural Language Processing (NLP) tasks. Their widespread adoption is primarily due to their proficiency in generating coherent text, comprehending complex linguistic structures, and providing contextually relevant responses. Numerous techniques have been developed to enhance the generative performance of LLMs and broaden their application areas. For instance, Chain-of-Thought~\citep{wei2022chain} illustrates LLMs' ability to formulate a distinct \textit{"thought process"} to tackle problems. WebGPT~\citep{nakano2021webgpt} employs LLMs to interact with web browsers, navigate web pages, and address complex queries effectively. Beyond conventional text generation tasks, Vision Large Language Models~\citep{wang2024visionllm, chi2024llama} extend the capabilities of traditional LLMs by incorporating visual comprehension, thus facilitating the convergence between textual and visual modalities. Minigpt-4~\citep{zhu2023minigpt} integrates a vision encoder with an LLM to augment visual understanding and multi-modal generation. Furthermore, Video-chatgpt~\citep{maaz2023video} introduces a model that merges video-adapted vision encoders with LLMs for intricate video analysis and dialogue generation. These models have significant utility in assisting with human tasks, including those in the medical field.



\section{Methodology}
\subsection{Problem Definition}
\label{sec:problem_definition}

As shown in Fig. \ref{fig:overview}, in the one-shot federated ensemble learning setting, there are \( N \) distributed clients (or parties) and a central model server, each client has its private dataset $D_i = \{(x_k, y_k)\}_{k=1}^{n_i}$, where  ${n_i}$ represents the size of the dataset for client $i$. The objective is to construct an optimal ensemble team $\mathcal{M} = \{M_i\}_{i=1}^{N}$, on dataset $\mathcal{D} = \{D_i\}_{i=1}^{N}$. This ensemble team comprises $N$ models, each independently trained by individual clients. In this setting, it is crucial to maintain all client data on their respective local devices, while only the models trained locally are sent to the server. In the one-shot setting, every client is allowed to communicate with the server only once. The objective is achieved by minimizing the loss of the ensemble team $\mathcal{M}$ on dataset $\mathcal{D}$, defined as follows:


\begin{equation}
    \mathop{\min} \mathbb{E}_{(x, y) \sim \mathcal{D}} \, \ell(f_{\mathcal{M}}(x), y),
\end{equation}

where $\ell$ is the loss function, $f_{\mathcal{M}}(x)$ denotes the prediction function based on $\mathcal{M}$, obtained as the result of ensemble learning, such as voting.

\subsection{Proposed Method: \fedmmeMethod}


\begin{figure*}[t]
    \centering
     \includegraphics[width=\linewidth]{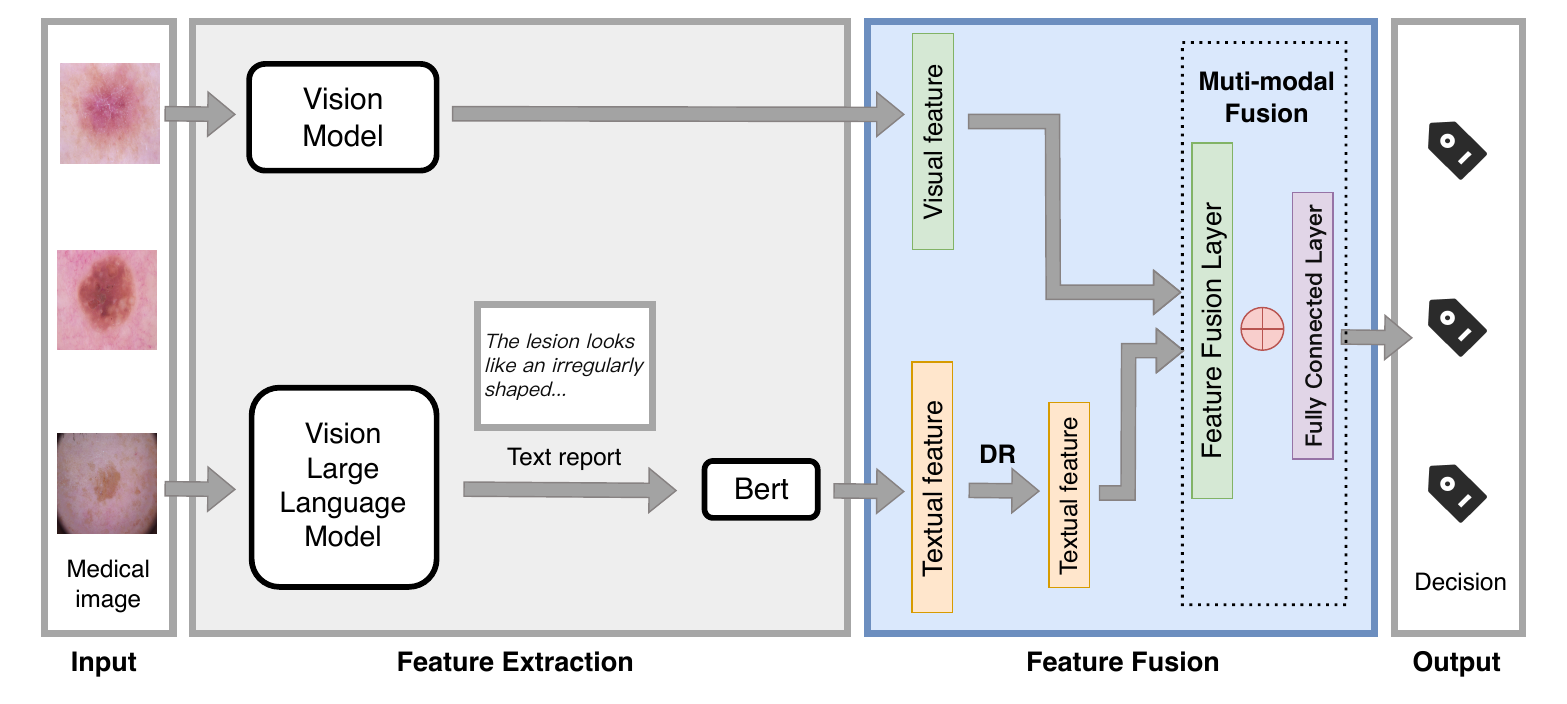}
     \caption{Overview of our proposed framework, \fedmmeMethod, which is structured into two principal phases: feature extraction and feature fusion.  In the feature extraction phase, visual features are obtained from images using conventional vision models, such as ResNet-18, and accompanying textual reports are produced via a sophisticated vision large language model. Textual features are subsequently extracted from these reports using a model designed for textual feature extraction, such as BERT. During the feature fusion phase, these visual features are combined with dimensionality-reduced textual features to create a cohesive feature representation. This combined representation is then processed through a fully connected layer, which enables classification.}

     \label{fig:method}
\end{figure*}


Numerous studies~\citep{huang2024knowledge, koga2025evaluating, zeiser2024multsurv} suggest that the integration of textual reports with image features significantly improves the performance of image classification compared to employing image features alone. Consequently, it is crucial to utilize multi-modal data when developing diagnostic tools in the healthcare industry. As depicted in Fig. \ref{fig:method}, we present a one-shot multi-modal federated ensemble framework, \fedmmeMethod, to address the issue outlined in Sec. \ref{sec:problem_definition}. In this framework, each client independently trains a multi-modal model on its local dataset utilizing a vision large language model, subsequently sending the trained model to a centralized server. On the server side, these models are aggregated to form an ensemble team. The ultimate decision is then determined through a voting mechanism involving the models within this ensemble team.

As shown in Fig. \ref{fig:method}, our method encompasses two primary phases: (1) \textit{Feature Extraction}: this phase involves extracting two types of features. Initially, high-dimensional visual features are extracted from medical images with conventional vision models. Subsequently, a vision large language model is employed to generate a report for the image, from which textual features are derived. (2) \textit{Feature Fusion}: during this phase, visual and textual features are concatenated to create a comprehensive feature layer used for multi-modal classification. To ensure that visual features maintain a dominant role in the prediction process—as the textual modality is intended to serve as an auxiliary—we perform dimensionality reduction on the textual features before integration, thus minimizing their impact. Finally, a fully connected layer is attached to this comprehensive feature layer to produce the final prediction results.

\begin{algorithm}[t]
    \caption{One-shot multi-modal federated ensemble framework}
    \label{alg:framework}
    \KwIn{Datasets $\mathcal{D} = \{D_i\}_{i=1}^{N}$ with $D_i = \{(x_k, y_k)\}_{k=1}^{n_i}$, learning rate $\eta$, number of local iterations $E_{local}$, batch size $B$, number of classes $Y$}
    \KwOut{The ensemble team $\mathcal{M} = \{M_i\}_{i=1}^{N}$} 
\begin{multicols}{2}
    \textbf{On server side:}\\
        
        \For{client $i= 1:N$ in parallel}{

        $M_i \leftarrow \mathbf{LocalTraining}(D_i)$

        }
        
    \textcolor{lightblue}{// The inference process for dataset $D=\{x_k\}_{k=1}^{n}$} \\
    \For{$k= 1:n$}{
    Initialize ${v} = \{0\}_{y=1}^{Y}$\\
    \For{$i= 1:N$}{
    $y_i=M_i(x_k)$
    \textcolor{lightblue}{// refer to Algorithm 2}\\
    $v[y_i] \leftarrow v[y_i]+1$
    }
    $y_{k}=\underset{y}{\argmax} \ v[y]$\\
    }
\columnbreak
\BlankLine
    \textbf{On client side:}\\
    $\mathbf{LocalTraining}(D_i)$: \\
        Initialize $M_i$ \\
        \For{each local epoch t from $1$ \KwTo $E_{local}$}{
        {
        Split $D_i$ into $\lceil n_i / B \rceil$ batches: $\{B_1, B_2, \dots, B_s\}$\\
        \For{batch $B_j$ in $\{B_1, B_2, \dots, B_s\}$}{
        $B_j$=$\{(x_{k}, y_{k})\}_{k=1}^{|B_j|}$\\
        $\hat{y}_{k} = M_i(x_k)$ for all $x_k \in B_j$\\
        $L(B_j) = \frac{1}{|B_j|} \sum_{y_k \in B_j} \ell(\hat{y}_{k}, y_{k})$\\
        $M_i \leftarrow M_i - \eta \nabla L(B_j)$\\
        }
        }
    }
    return $M_i$ to the server
\end{multicols}
\end{algorithm}

\begin{algorithm}[t]
    \caption{\fedmmeMethod}
    \label{alg:method}
    \KwIn{Original image $x$}
    \KwOut{The prediction result $y_{pred}$} 
    \textcolor{lightblue}{1. Extract the image features with a traditional vision model}\\
    $f_{visual} = VisionModel(x)$ \\
    \textcolor{lightblue}{2. Extract the textual features with a vision large language model}\\
    $r = VisionLLM(x)$    \textcolor{lightblue}{// r: the report of images}\\
    $f_{textual} = BERT(r)$\\
    \textcolor{lightblue}{3. Integrate the extracted visual and textual features}\\
    
    $f_{textual}^{*} = DR(f_{textual}) $\textcolor{lightblue}{// \textit{DR}($\cdot$): Dimensionality Reduction}\\
    
    $f_{combine} = f_{visual} + f_{textual}^{*}$\\

    \textcolor{lightblue}{4. Get the prediction result}\\
    
    $y_{pred} = FC(f_{combine}) $\textcolor{lightblue}{// \textit{FC}($\cdot$): fully connected layer}\\
    return $y_{pred}$

\end{algorithm}

\subsubsection{Feature Extraction}



\textbf{\textit{Visual Feature Extraction}}: We utilize a conventional vision model, such as the ResNet-18~\citep{he2016deep} model with the final layer removed, to extract visual features from medical images. This approach is extensively employed in image processing tasks. For a given medical image $x$, the extracted visual features can be represented as:

\begin{equation}
f_{visual} = VisionModel(x),
\end{equation}
where $f_{visual} \in \mathbb{R}^{d_1}$ denotes the feature vector extracted by the vision model, and $d_1$ is the dimension of the feature vector. The vision model is pre-trained to accurately capture critical features in medical images, thereby ensuring optimal performance in subsequent tasks. This model will also be incorporated into the subsequent multi-modal training process to further enhance the effectiveness of our ensemble team.


\textbf{\textit{Text Report Generation}}: Our method seeks to enhance the diagnostic accuracy of medical images by incorporating textual features, and vision large language models have been demonstrated to significantly enhance feature richness by interpreting the context within images~\citep{zhou2024automated, ding2024hia}. These models function by comprehending and generating textual descriptions from visual data, thereby bridging the gap between visual perception and linguistic comprehension. By leveraging these models, our system can identify subtle details often overlooked by traditional image-only methods. However, employing solely a vision large language model for the classification of medical images generally results in suboptimal outcomes. For example, when evaluated under the identical experimental conditions described in our study when the Dirichlet distribution $\alpha=$0.6, the Llama-3.2-8B-Vision-Instruct~\citep{chi2024llama} model only achieves a classification accuracy of 29\% on the Blood~\citep{yang2023medmnist}  dataset. This performance is significantly inferior to the approximate 80\% accuracy achieved by purely visual classification models such as ResNet-18. Consequently, it is inadvisable to rely exclusively on vision large language models for image classification tasks. Instead, these models should be incorporated as ancillary training resources within a multi-modal framework to augment the efficacy of vision models. Therefore, after the extraction of image features, we utilize a vision large language model, such as the Llama-3.2-11B-Vision-Instruct model, to generate textual reports for the medical image:

\begin{equation}
r = VisionLLM(x),
\end{equation}

where $r$ denotes the generated report for the medical image $x$. 

\textbf{\textit{Textual Feature Extraction}}: The text generated by the large vision language model is presented in textual format rather than as visual features like those extracted by the vision model. Consequently, we cannot directly merge it with the visual features for input into a multi-modal model. To address this, we utilize a pre-trained BERT~\citep{devlin2018bert, onan2023hierarchical} model to derive textual features from the report. BERT has exhibited an exceptional ability to grasp the context and semantics of text across a multitude of natural language processing applications. The textual features produced by BERT are in the form of embeddings—dense representations that encapsulate various dimensions and interrelationships within the textual data. These embeddings can then be concatenated with the visual features for prediction purposes. The process of extracting textual features can be depicted as follows:



\begin{equation}
f_{textual} = BERT(r),
\end{equation}
where \(f_{textual} \in \mathbb{R}^{d_2}\) denotes the feature vector extracted by BERT, $r$ is the textual description generated by the vision large language model. $d_2$ represents the dimension of the textual feature vector. 

\subsubsection{Feature Fusion}
\textbf{\textit{Dimensionality Reduction}}: When merging textual and visual features during training, it is crucial to ensure that the textual elements do not overpower or displace visual information as the main modality. If textual features dominate, it could significantly compromise the generalization capability of the model~\citep{rahman2020integrating, fei2022towards}. For instance, in employing the ResNet-18 model for visual features extraction and the BERT model for textual features, visual features from ResNet are noted to have a dimensionality of 512, in contrast to the 768 of BERT. This greater dimensionality of textual features might inadvertently suppress the visual modality, negating our original purpose.

Therefore, to prevent textual features from overshadowing visual features, we apply a dimension-reduction strategy on the textual features. This reduction aims to curtail their influence on the overall effectiveness of the model, thereby ensuring that visual features maintain a more prominent role in driving the model’s predictions. The dimensionality reduction process is outlined as follows:

\begin{equation}
f_{textual}^{*} = DR(f_{textual}),
\end{equation}
where $f_{textual}^{*}$ represents the textual features after the dimensionality reduction process, denoted as $DR(\cdot)$. \\


\textbf{\textit{Feature Concatenation}}: After reducing the dimensionality of textual features, we concatenate them with visual features to form a unified feature representation for training our multi-modal model. This integration harnesses the strengths of each modality, thereby enhancing both the robustness and accuracy of the model predictions. The combined feature vector is represented as follows:

\begin{equation}
f_{combine} = f_{visual} + f_{textual}^{*},
\end{equation}

where $f_{combine}$ denotes the new feature vector created by concatenating the visual features with the dimensionality-reduced textual features.

Finally, we attach a fully connected layer to the integrated feature vector to produce the final prediction result for the corresponding image.

\begin{equation}
y_{pred} = FC(f_{combine}),
\end{equation}

where $FC(\cdot)$ represents the application of a fully connected layer to the concatenated features.


Alg. \ref{alg:framework} presents a comprehensive overview of the federated ensemble learning framework and the inference mechanisms using our multi-modal model, where $M_i$ denotes our designed multi-modal model on client $i$. Meanwhile, Alg. \ref{alg:method} details the procedure of our \fedmmeMethod\ framework for generating and handling multi-modal data features. After obtaining the entire ensemble team $\mathcal{M}$, ensemble learning techniques can be applied to derive the final prediction outcomes, for example, through the utilization of voting methods.

\section{Experiments}


\subsection{Experiments Setup}
\textbf{Datasets.} We selected four representative medical datasets to evaluate our framework: Blood and Derma from the MedMNIST dataset~\citep{yang2023medmnist}, RSNA~\citep{rsna2019rsna}, and Diabetic~\citep{diabetic-retinopathy-detection}. These datasets cover a range of medical application scenarios. To closely mimic real-world conditions, we conducted extensive experiments across various Non-IID settings—a prevalent challenge in federated learning—to comprehensively evaluate the efficacy of our method. Specifically, we employed a Dirichlet distribution to allocate datasets among $N=5$ clients using concentration parameters $\alpha = 0.6, 0.3,$ and $0.1$. This approach enabled us to explore different levels of data heterogeneity and rigorously test the robustness of our methodology. 

\textbf{Settings.} In our experiments, the vision model employs the ResNet-18~\citep{he2016deep} architecture to extract visual features from original medical images. Local client models were trained for 100 epochs with SGD optimizer using learning rate (LR) 1e-3 and batch size 128. Additionally, we utilize the Llama-3.2-11B-Vision-Instruct~\citep{chi2024llama} model to generate textual reports from medical images and employ a BERT~\citep{devlin2018bert} model to extract textual features from these reports. Finally, we utilize the equal-weight voting~\citep{raza2019improving} strategy to implement the ultimate ensemble learning process.

All our experiments are running on a single machine with 1TB RAM and 256 cores AMD EPYC 7742 64-Core Processor @ 3.4GHz CPU. The GPU we used is NVIDIA A100 SXM4 with 40GB memory. 
The environment settings are: Python 3.9.12, PyTorch 1.12.1 with CUDA 11.6
on Ubuntu 20.04.4 LTS.

All the experimental results are the average over three trials.

 


\subsection{Baselines}

We compared our proposed method with four established baseline methods: \textbf{FedAvg}~\citep{mcmahan2017communication}, a well-known federated learning algorithm that conducts multi-round optimization of a distributed model by performing local training on clients and using weighted averaging for aggregation on the server. For consistency with our method, we adapted FedAvg to a \textit{one-shot} setting. \textbf{DENSE}~\citep{zhang2022dense} represents a one-shot federated learning framework relying on a central server to disseminate a global model through knowledge distillation. \textbf{DAFL}~\citep{chen2019data} is another one-shot federated learning strategy, in which a teacher model is built from multiple client models using knowledge distillation to develop a global model. \textbf{FedISCA}~\citep{kang2023one}, the state-of-the-art one-shot federated learning framework, is specifically designed for medical applications and addresses data heterogeneity through synthetic data generation and client model adaptation with the help of knowledge distillation. Additionally, we incorporated \textbf{FedEnsemble} into our comparison. This method entails each local client employing a single modal vision model, such as Resnet-18 in our experiments, to facilitate ensemble learning on a central server.


\subsection{Performance Analysis}

Table~\ref{table:main_results} presents the accuracy results for various datasets across different data partitions and methods. As we can see, our method consistently outperforms all other baselines across all data partitions and datasets. Notably, it surpasses existing one-shot federated learning approaches by more than 17.5\% in accuracy on the RSNA dataset when applying a Dirichlet distribution with ($\alpha=0.3$). This significant improvement underscores the efficacy of our framework in one-shot federated ensemble learning within medical contexts. Furthermore, although the accuracy of all methods tends to decrease as the degree of non-IID increases, our method persistently exhibits superior performance, thereby illustrating its robustness.



It is essential to acknowledge that numerous baseline methods encounter difficulties in achieving high performance on the Diabetic dataset, which stands as the largest dataset embodying real-world data. For instance, some approaches, such as DAFL, only manage to reach accuracy levels akin to that of random guessing, thereby underscoring their inadequacies in handling real-world scenarios. In contrast, our proposed method exhibits strong and consistent performance, significantly outperforming other techniques. More precisely, when the Dirichlet parameter is set to $\alpha=0.3$, our method attains an accuracy of 31.93\%, which not only surpasses DAFL by 11\% but also exceeds the second-best method, FedISCA, by over 3\%.


\begin{table}[t]
  \centering
  \label{table:main_results}
  \caption{Test accuracy comparison for different datasets on various data partitions and methods. \textbf{Bold} indicates the best accuracy among one-shot FL methods.}
\scalebox{0.8}{
\begin{tabular}{c|cccc|cccc|cccc}
\myhline
 & \multicolumn{4}{c|}{Dirichlet ($\alpha$=0.6)} & \multicolumn{4}{c|}{Dirichlet ($\alpha$=0.3)} & \multicolumn{4}{c}{Dirichlet ($\alpha$=0.1)} \\
\hline
Dataset  & Blood & Derma & RSNA  & Diabetic & Blood & Derma & RSNA  & Diabetic & Blood & Derma & RSNA  & Diabetic \\
\hline
    FedAvg & 19.48 & 69.23 & 70.39 & 20.01 & 30.51 & 11.02 & 69.31 & 19.99 & 19.81 & 66.88 & 68.55 & 19.98 \\

    DAFL  & 17.13 & 63.33 & 50.44 & 20.03 & 16.03 & 13.64 & 48.88 & 20.04 & 15.11 & 12.58 & 46.69 & 20.01 \\

    DENSE & 34.52 & 64.78 & 55.04 & 23.31 & 30.17 & 12.78 & 51.08 & 23.22 & 28.87 & 11.37 & 47.79 & 21.32 \\

    FedISCA & 53.61 & 53.86 & 70.59 & 26.98 & 48.93 & 16.11 & 70.39 & 28.79 & 53.61 & 53.86 & 69.11 & 20.97 \\

    \hline

    FedEnsemble & 84.33 & 67.23 & 83.46 & 24.85 & 71.03 & 66.13 & 70.42 & 25.48 & 54.92 & 67.13 & 68.76 & 20.54 \\

    \textbf{\fedmmeMethod} & \textbf{87.72} & \textbf{71.27} & \textbf{85.49} & \textbf{29.81} & \textbf{80.12} & \textbf{71.27} & \textbf{87.93} & \textbf{31.93} & \textbf{68.11} & \textbf{69.13} & \textbf{71.38} & \textbf{21.55} \\
\myhline
\end{tabular}%
}
  \label{table:main_results}%
\end{table}%

Meanwhile, we conducted experiments with varying numbers of clients to assess the scalability of our method. Table~\ref{table:num_users} demonstrates that our approach consistently outperforms other baselines across all scenarios involving different numbers of clients on the Blood dataset under a Dirichlet distribution ($\alpha = 0.3$). This underscores its feasibility for large-scale federated learning environments and attests to its robustness. Additionally, we observed an improvement in the performance of most methods as the number of models increased with more clients. This enhancement can be attributed to the increased diversity within the group of models, which, in turn, leads to superior ensemble or aggregation performance~\citep{wu2021boosting, wang2023data}, particularly compared to scenarios with fewer clients.


\begin{table}[t]
  \centering
  \label{table:num_users}
  \caption{Performance comparison on the Blood dataset with the different number of clients, where Dirichlet parameter $\alpha = 0.3$.}
    \begin{tabular}{c|c|c|c|c|c|c}
    \myhline
    Number of Users & FedAvg & DAFL  & DENSE & FedISCA & FedEnsemble & \fedmmeMethod \\
    \hline
    5     & 30.51 & 16.03 & 30.17 & 48.93 & 71.03 & \textbf{80.12} \\
    10    & 20.22 & 17.11 & 31.28 & 49.99 & 77.17 & \textbf{82.72} \\
    20    & 20.11 & 18.29 & 33.25 & 52.47 & 78.16 & \textbf{86.23} \\
    \myhline
    \end{tabular}%
  \label{table:num_users}
\end{table}%

\begin{figure*}[t]
    \centering
     \includegraphics[width=0.7\linewidth]{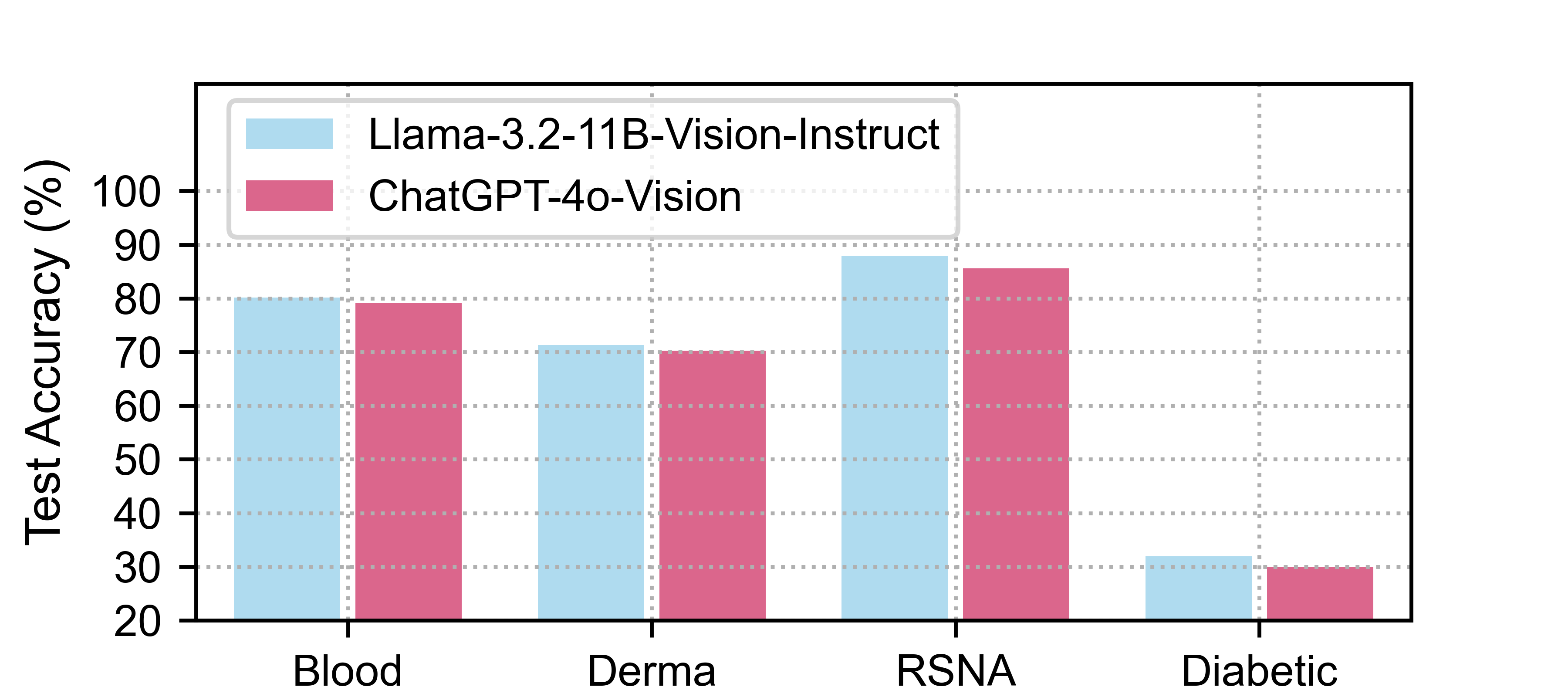}
     \caption{Test Accuracy comparison of \fedmmeMethod\ with different vision large language models, where Dirichlet parameter $\alpha = 0.3$.}

     \label{fig:diff_llm}
\end{figure*}

\subsection{Ablation Studies}
\subsubsection{Comparative analysis of Vision Large Language Model}

In the process of generating textual reports from medical images, we hypothesized that diverse vision large language models might output varying reports when interpreting the same image, potentially impacting the efficacy of our multi-modal training approach. Consequently, we assessed the performance of two vision large language models, Llama-3.2-11B-Vision-Instruct and ChatGPT-4o-Vision, across multiple datasets characterized by a Dirichlet distribution with ($\alpha=$0.3). Fig. \ref{fig:diff_llm} shows that the Llama-3.2-11B-Vision-Instruct model consistently outperformed the ChatGPT-4o-Vision model in all four datasets. This is because ChatGPT-4o-Vision often refrained from responding to domain-specific queries, typically responding with, \textit{"I'm sorry, I can't assist with identifying the type of blood cell from this image".} However, when we instructed the ChatGPT-4o-Vision model to focus on describing the features of the images, rather than directly categorizing them through tailored prompts, both models achieved comparable performance. This underscores the adaptability of our method across different types of vision large language models.


\subsubsection{Effects of different textual feature size}



\begin{figure*}[t]
    \centering
     \includegraphics[width=0.7\linewidth]{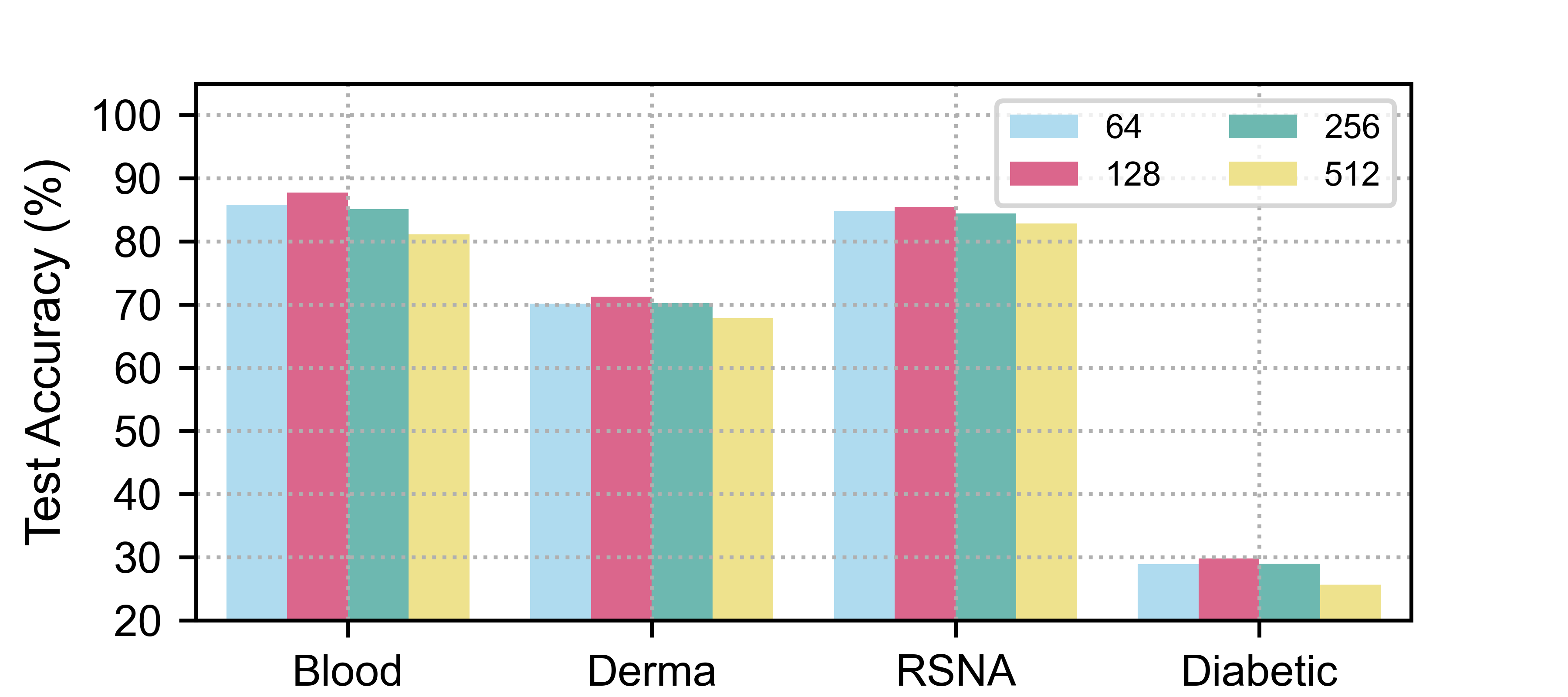}
     \caption{Effect of various textual feature sizes across different datasets, where Dirichlet parameter $\alpha = 0.6$.}

     \label{fig:diff_feature_size}
\end{figure*}

To assess the impact of dimension-reduced textual features on our framework, we conducted experiments in which we reduced the original textual features to various dimensions. Fig. \ref{fig:diff_feature_size} illustrates that when the textual feature size remains below 512, the impact on performance is negligible, with optimal results at a dimension of 128. This indicates that our method maintains robustness across different sizes of textual features. However, performance deteriorates markedly as the textual feature size increases to 512—the point at which it equals the visual feature size. This decline is attributed to the excessive size of the textual features, which nearly supplants the primary visual modality and consequently degrades overall performance. Thus, maintaining the dimensionality reduction component is crucial for the efficacy of the multi-modal model.

\subsubsection{Training convergence analysis}



\begin{figure*}[t]
    \centering
    \begin{minipage}[b]{0.45\textwidth}
        \centering
        \includegraphics[width=\linewidth]{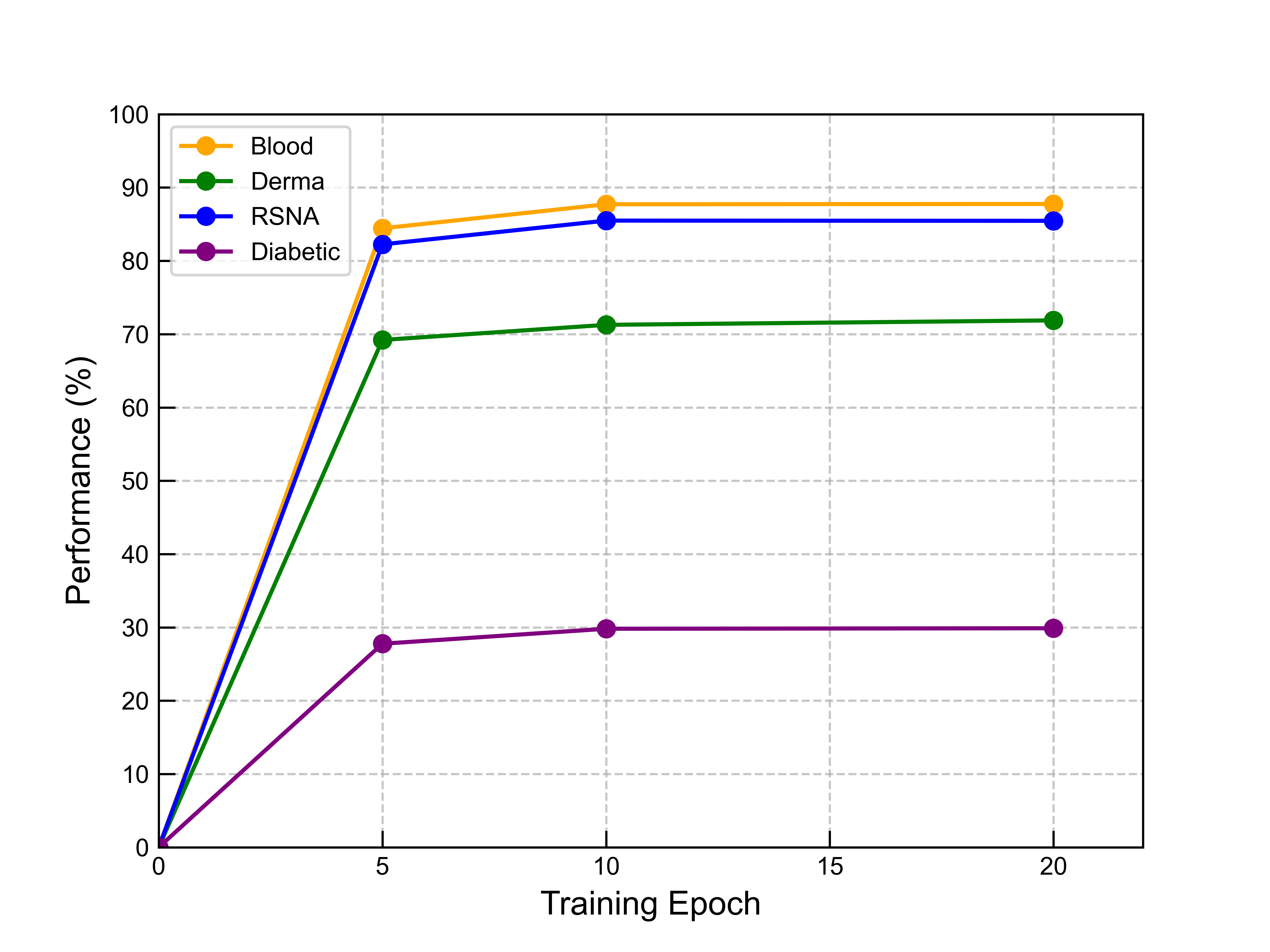}
        \caption{Training convergence analysis by various training epochs, where Dirichlet parameter $\alpha = 0.6$.}
        \label{fig:train_epoch}
    \end{minipage}
    \hspace{0.05\textwidth} 
    \begin{minipage}[b]{0.4\textwidth}
        \centering
        \includegraphics[width=0.76\linewidth]{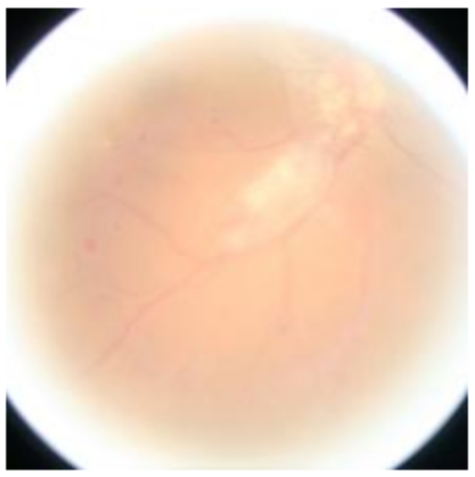}
        \caption{Retinal image with several features that are typical of diabetic retinopathy (DR).}
        \label{fig:case}
    \end{minipage}
\end{figure*}

Fig. \ref{fig:train_epoch} illustrates the performance of our method across varying numbers of local training epochs for all evaluated datasets. Notably, our model demonstrates substantial accuracy improvements after just 10 rounds of local training, indicating rapid convergence and sustained high performance with fewer iterations. Additionally, the results underscore the method's robustness, as it maintains consistent performance regardless of changes in the number of local training epochs. This consistency highlights the method's adaptability and reliability in diverse scenarios.



\begin{figure*}[t]
    \centering
     \includegraphics[width=1\linewidth]{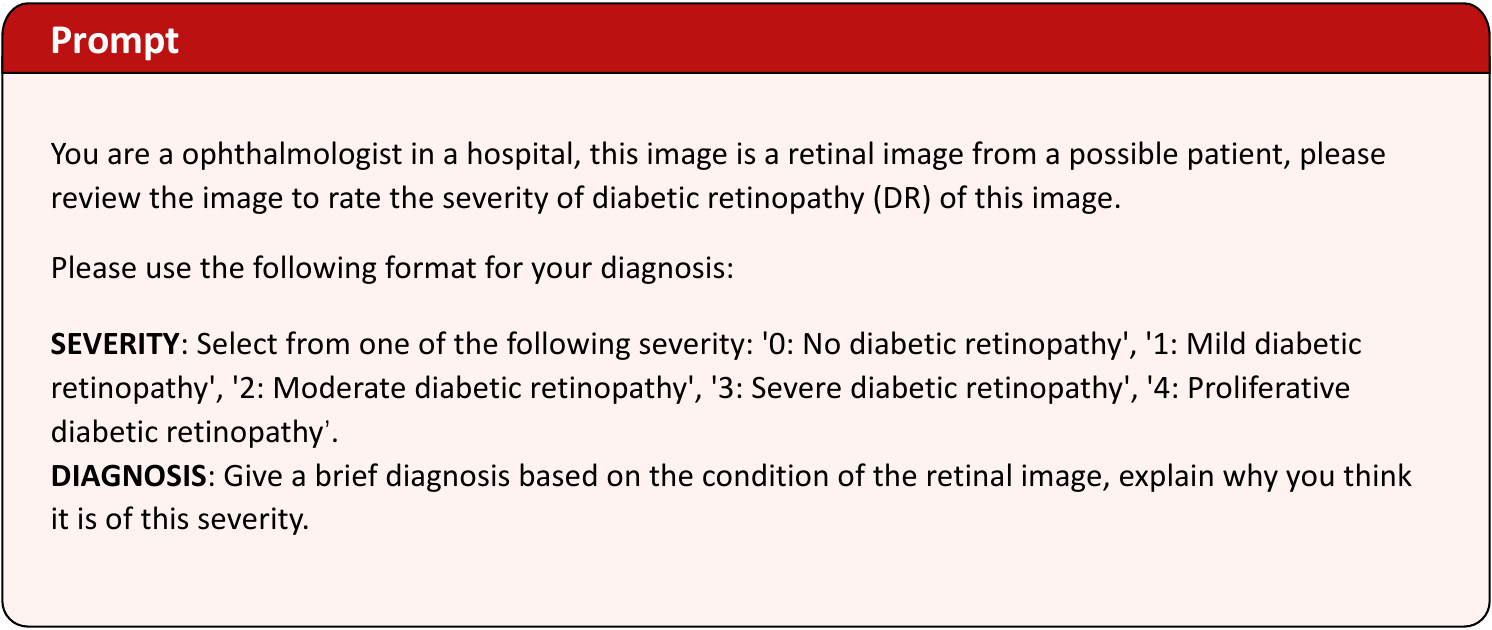}

    \caption{The prompt used for the Llama-3.2-11B-Vision-Instruct model to recognize the type of the medical image shown in Fig. \ref{fig:case}.}
     \label{fig:prompt}
\end{figure*}

\subsection{Case Study}



In this section, we present a case study to illustrate the effectiveness of the vision large language model incorporated into our framework. Fig. \ref{fig:case} is derived from the Diabetic Retinopathy dataset, primarily utilized to evaluate the severity of Diabetic Retinopathy (DR), a common complication of diabetes. Accurate assessment of DR severity is crucial for devising effective treatment plans for patients.

\begin{figure*}[t]
    \centering
     \includegraphics[width=1\linewidth]{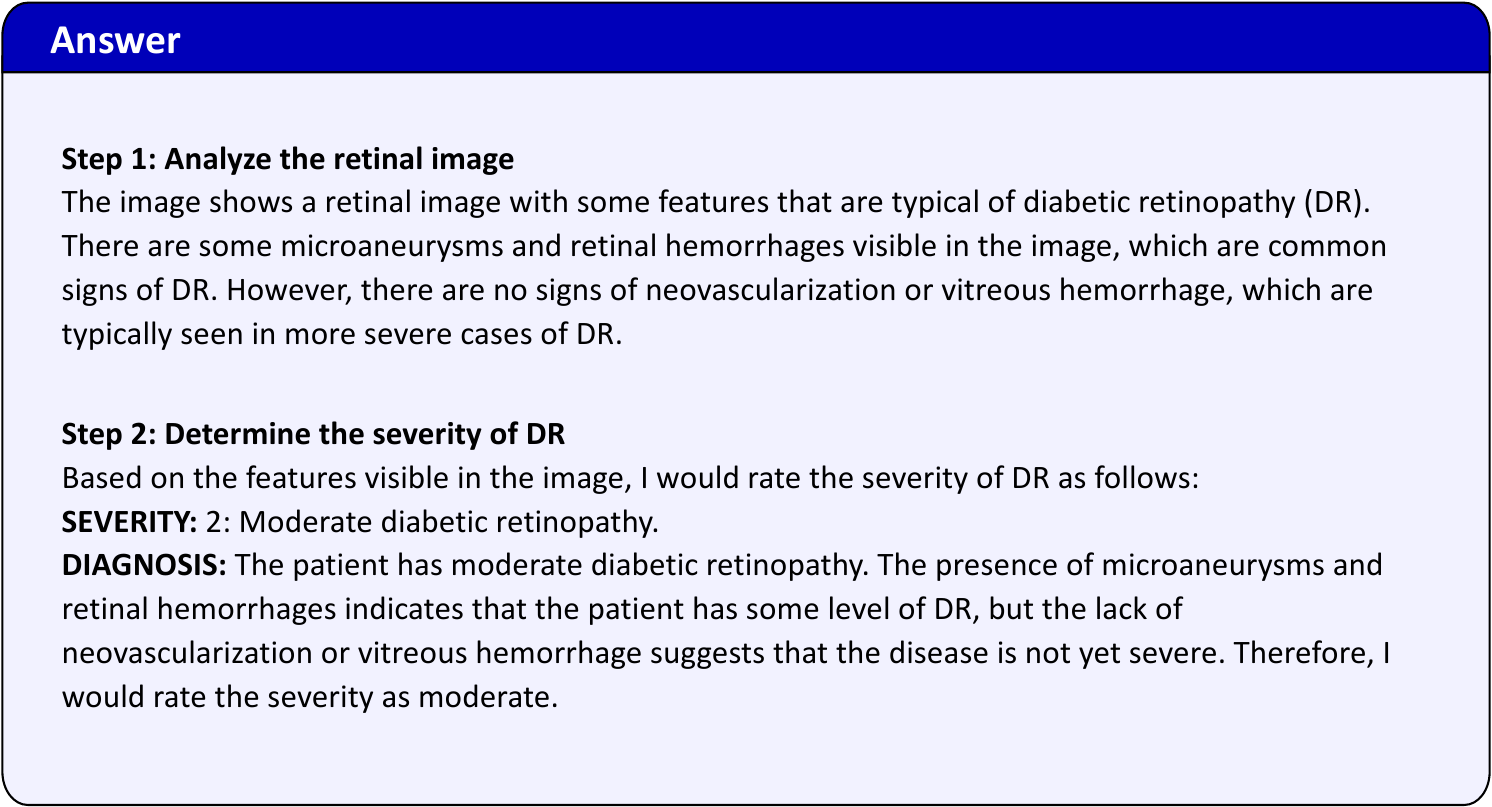}

    \caption{The output generated by the Llama-3.2-11B-Vision-Instruct model in response to the prompt presented in Fig. \ref{fig:prompt} and the medical image depicted in Fig. \ref{fig:case}.}
     \label{fig:answer}
\end{figure*}

Fig. \ref{fig:prompt} and Fig. \ref{fig:answer} illustrate the prompt and response utilized by the vision large language model, Llama-3.2-11B-Vision-Instruct, to categorize and analyze the medical image depicted in Fig. \ref{fig:case}. In this case, despite the Llama-3.2-11B-Vision-Instruct model incorrectly classifying the severity as \textit{"Moderate diabetic retinopathy"} rather than the correct severity \textit{"Proliferative diabetic retinopathy"}, the text report still provides valuable insights. It successfully identifies specific retinal features, including hemorrhages and microaneurysms. These detailed descriptions of the images significantly contribute to improving the accuracy of predictions made by multi-modal models. Therefore, our multi-modal training framework, augmented by the vision large language model, demonstrates to be a significant tool in the realm of medical diagnostic applications.


\section{Conclusion}

In this paper, we present a novel multi-modal one-shot federated ensemble learning framework for medical image analysis. By incorporating both visual and textual features with the help of large vision language models, our method not only addresses the challenges posed by data privacy but also improves the robustness and accuracy of medical diagnostics. We demonstrated the effectiveness of our framework across multiple datasets, showing its superior performance in analyzing diverse medical images compared to traditional federated learning methods. To the best of our knowledge, this is the first work to systematically explore the one-shot federated ensemble learning for training multi-modal models.

\bibliography{main}

\begin{thebibliography}{68}
\providecommand{\natexlab}[1]{#1}
\providecommand{\url}[1]{\texttt{#1}}
\expandafter\ifx\csname urlstyle\endcsname\relax
  \providecommand{\doi}[1]{doi: #1}\else
  \providecommand{\doi}{doi: \begingroup \urlstyle{rm}\Url}\fi

\bibitem[Abaoud et~al.(2023)Abaoud, Almuqrin, and Khan]{abaoud2023advancing}
Mohammed Abaoud, Muqrin~A Almuqrin, and Mohammad~Faisal Khan.
\newblock Advancing federated learning through novel mechanism for privacy preservation in healthcare applications.
\newblock \emph{IEEE Access}, 11:\penalty0 83562--83579, 2023.

\bibitem[Alam et~al.(2023)Alam, Ahmed, Hossain, Emo, Bidhan, Reza, Alam, Hassan, Pupo, and Fortino]{alam2023federated}
Md~Mustakin Alam, Tanjim Ahmed, Meraz Hossain, Mehedi~Hasan Emo, Md~Kausar~Islam Bidhan, Md~Tanzim Reza, Md~Golam~Rabiul Alam, Mohammad~Mehedi Hassan, Francesco Pupo, and Giancarlo Fortino.
\newblock Federated ensemble-learning for transport mode detection in vehicular edge network.
\newblock \emph{Future Generation Computer Systems}, 149:\penalty0 89--104, 2023.

\bibitem[Balkus et~al.(2022)Balkus, Wang, Cornet, Mahabal, Ngo, and Fang]{balkus2022survey}
Salvador~V Balkus, Honggang Wang, Brian~D Cornet, Chinmay Mahabal, Hieu Ngo, and Hua Fang.
\newblock A survey of collaborative machine learning using 5g vehicular communications.
\newblock \emph{IEEE Communications Surveys \& Tutorials}, 24\penalty0 (2):\penalty0 1280--1303, 2022.

\bibitem[Breiman(1996)]{breiman1996bagging}
Leo Breiman.
\newblock Bagging predictors.
\newblock \emph{Machine learning}, 24:\penalty0 123--140, 1996.

\bibitem[Brown et~al.(2020)Brown, Mann, Ryder, Subbiah, Kaplan, Dhariwal, Neelakantan, Shyam, Sastry, Askell, et~al.]{brown2020language}
Tom Brown, Benjamin Mann, Nick Ryder, Melanie Subbiah, Jared~D Kaplan, Prafulla Dhariwal, Arvind Neelakantan, Pranav Shyam, Girish Sastry, Amanda Askell, et~al.
\newblock Language models are few-shot learners.
\newblock \emph{Advances in neural information processing systems}, 33:\penalty0 1877--1901, 2020.

\bibitem[Caldas et~al.(2018)Caldas, Kone{\v{c}}ny, McMahan, and Talwalkar]{caldas2018expanding}
Sebastian Caldas, Jakub Kone{\v{c}}ny, H~Brendan McMahan, and Ameet Talwalkar.
\newblock Expanding the reach of federated learning by reducing client resource requirements.
\newblock \emph{arXiv preprint arXiv:1812.07210}, 2018.

\bibitem[Chakravarty et~al.(2021)Chakravarty, Kar, Sethuraman, and Sheet]{chakravarty2021federated}
Arunava Chakravarty, Avik Kar, Ramanathan Sethuraman, and Debdoot Sheet.
\newblock Federated learning for site aware chest radiograph screening.
\newblock In \emph{2021 IEEE 18th International Symposium on Biomedical Imaging (ISBI)}, pp.\  1077--1081. IEEE, 2021.

\bibitem[Chen et~al.(2019)Chen, Wang, Xu, Yang, Liu, Shi, Xu, Xu, and Tian]{chen2019data}
Hanting Chen, Yunhe Wang, Chang Xu, Zhaohui Yang, Chuanjian Liu, Boxin Shi, Chunjing Xu, Chao Xu, and Qi~Tian.
\newblock Data-free learning of student networks.
\newblock In \emph{Proceedings of the IEEE/CVF international conference on computer vision}, pp.\  3514--3522, 2019.

\bibitem[Chi et~al.(2024)Chi, Karn, Zhan, Smith, Rando, Zhang, Plawiak, Coudert, Upasani, and Pasupuleti]{chi2024llama}
Jianfeng Chi, Ujjwal Karn, Hongyuan Zhan, Eric Smith, Javier Rando, Yiming Zhang, Kate Plawiak, Zacharie~Delpierre Coudert, Kartikeya Upasani, and Mahesh Pasupuleti.
\newblock Llama guard 3 vision: Safeguarding human-ai image understanding conversations.
\newblock \emph{arXiv preprint arXiv:2411.10414}, 2024.

\bibitem[Chowdhery et~al.(2023)Chowdhery, Narang, Devlin, Bosma, Mishra, Roberts, Barham, Chung, Sutton, Gehrmann, et~al.]{chowdhery2023palm}
Aakanksha Chowdhery, Sharan Narang, Jacob Devlin, Maarten Bosma, Gaurav Mishra, Adam Roberts, Paul Barham, Hyung~Won Chung, Charles Sutton, Sebastian Gehrmann, et~al.
\newblock Palm: Scaling language modeling with pathways.
\newblock \emph{Journal of Machine Learning Research}, 24\penalty0 (240):\penalty0 1--113, 2023.

\bibitem[Cui et~al.(2021)Cui, Yin, Wang, Li, and Wang]{cui2021stacking}
Shaoze Cui, Yunqiang Yin, Dujuan Wang, Zhiwu Li, and Yanzhang Wang.
\newblock A stacking-based ensemble learning method for earthquake casualty prediction.
\newblock \emph{Applied Soft Computing}, 101:\penalty0 107038, 2021.

\bibitem[Dennis et~al.(2021)Dennis, Li, and Smith]{dennis2021heterogeneity}
Don~Kurian Dennis, Tian Li, and Virginia Smith.
\newblock Heterogeneity for the win: One-shot federated clustering.
\newblock In \emph{International Conference on Machine Learning}, pp.\  2611--2620. PMLR, 2021.

\bibitem[Devlin(2018)]{devlin2018bert}
Jacob Devlin.
\newblock Bert: Pre-training of deep bidirectional transformers for language understanding.
\newblock \emph{arXiv preprint arXiv:1810.04805}, 2018.

\bibitem[Ding et~al.(2024)Ding, Chu, Pi, Wang, and Li]{ding2024hia}
Xinpeng Ding, Yongqiang Chu, Renjie Pi, Hualiang Wang, and Xiaomeng Li.
\newblock Hia: Towards chinese multimodal llms for comparative high-resolution joint diagnosis.
\newblock In \emph{International Conference on Medical Image Computing and Computer-Assisted Intervention}, pp.\  575--586. Springer, 2024.

\bibitem[Drainakis et~al.(2020)Drainakis, Katsaros, Pantazopoulos, Sourlas, and Amditis]{drainakis2020federated}
Georgios Drainakis, Konstantinos~V Katsaros, Panagiotis Pantazopoulos, Vasilis Sourlas, and Angelos Amditis.
\newblock Federated vs. centralized machine learning under privacy-elastic users: A comparative analysis.
\newblock In \emph{2020 IEEE 19th International Symposium on Network Computing and Applications (NCA)}, pp.\  1--8. IEEE, 2020.

\bibitem[Dugas et~al.(2015)Dugas, Jared, Jorge, and Cukierski]{diabetic-retinopathy-detection}
Emma Dugas, Jared, Jorge, and Will Cukierski.
\newblock Diabetic retinopathy detection.
\newblock \url{https://kaggle.com/competitions/diabetic-retinopathy-detection}, 2015.
\newblock Kaggle.

\bibitem[Fei et~al.(2022)Fei, Lu, Gao, Yang, Huo, Wen, Lu, Song, Gao, Xiang, et~al.]{fei2022towards}
Nanyi Fei, Zhiwu Lu, Yizhao Gao, Guoxing Yang, Yuqi Huo, Jingyuan Wen, Haoyu Lu, Ruihua Song, Xin Gao, Tao Xiang, et~al.
\newblock Towards artificial general intelligence via a multimodal foundation model.
\newblock \emph{Nature Communications}, 13\penalty0 (1):\penalty0 3094, 2022.

\bibitem[Gadekallu et~al.(2021)Gadekallu, Pham, Huynh-The, Bhattacharya, Maddikunta, and Liyanage]{gadekallu2021federated}
Thippa~Reddy Gadekallu, Quoc-Viet Pham, Thien Huynh-The, Sweta Bhattacharya, Praveen Kumar~Reddy Maddikunta, and Madhusanka Liyanage.
\newblock Federated learning for big data: A survey on opportunities, applications, and future directions.
\newblock \emph{arXiv preprint arXiv:2110.04160}, 2021.

\bibitem[Ge et~al.(2024)Ge, Huang, Zhou, Li, Wang, Tang, and Zhuang]{ge2024worldgpt}
Zhiqi Ge, Hongzhe Huang, Mingze Zhou, Juncheng Li, Guoming Wang, Siliang Tang, and Yueting Zhuang.
\newblock Worldgpt: Empowering llm as multimodal world model.
\newblock In \emph{Proceedings of the 32nd ACM International Conference on Multimedia}, pp.\  7346--7355, 2024.

\bibitem[Geyer et~al.(2017)Geyer, Klein, and Nabi]{geyer2017differentially}
Robin~C Geyer, Tassilo Klein, and Moin Nabi.
\newblock Differentially private federated learning: A client level perspective.
\newblock \emph{arXiv preprint arXiv:1712.07557}, 2017.

\bibitem[Ghosh et~al.(2024)Ghosh, Acharya, Jain, Saha, Chadha, and Sinha]{ghosh2024clipsyntel}
Akash Ghosh, Arkadeep Acharya, Raghav Jain, Sriparna Saha, Aman Chadha, and Setu Sinha.
\newblock Clipsyntel: clip and llm synergy for multimodal question summarization in healthcare.
\newblock In \emph{Proceedings of the AAAI Conference on Artificial Intelligence}, volume~38, pp.\  22031--22039, 2024.

\bibitem[Guha et~al.(2018)Guha, Talwalkar, and Smith]{guha2018one}
Neel Guha, Ameet Talwalkar, and Virginia Smith.
\newblock One-shot federated learning.
\newblock In \emph{NeurIPS 2018 Workshop on Machine Learning on the Phone and other Consumer Devices}, 2018.

\bibitem[Guha et~al.(2019)Guha, Talwalkar, and Smith]{guha2019one}
Neel Guha, Ameet Talwalkar, and Virginia Smith.
\newblock One-shot federated learning.
\newblock \emph{arXiv preprint arXiv:1902.11175}, 2019.

\bibitem[He et~al.(2016)He, Zhang, Ren, and Sun]{he2016deep}
Kaiming He, Xiangyu Zhang, Shaoqing Ren, and Jian Sun.
\newblock Deep residual learning for image recognition.
\newblock In \emph{Proceedings of the IEEE conference on computer vision and pattern recognition}, pp.\  770--778, 2016.

\bibitem[Hoffmann et~al.(2022)Hoffmann, Borgeaud, Mensch, Buchatskaya, Cai, Rutherford, Casas, Hendricks, Welbl, Clark, et~al.]{hoffmann2022training}
Jordan Hoffmann, Sebastian Borgeaud, Arthur Mensch, Elena Buchatskaya, Trevor Cai, Eliza Rutherford, Diego de~Las Casas, Lisa~Anne Hendricks, Johannes Welbl, Aidan Clark, et~al.
\newblock Training compute-optimal large language models.
\newblock \emph{arXiv preprint arXiv:2203.15556}, 2022.

\bibitem[Hu et~al.(2024)Hu, Xu, Li, Li, Chen, and Tu]{hu2024bliva}
Wenbo Hu, Yifan Xu, Yi~Li, Weiyue Li, Zeyuan Chen, and Zhuowen Tu.
\newblock Bliva: A simple multimodal llm for better handling of text-rich visual questions.
\newblock In \emph{Proceedings of the AAAI Conference on Artificial Intelligence}, volume~38, pp.\  2256--2264, 2024.

\bibitem[Huang et~al.(2024)Huang, Cao, Jia, Li, Tang, and Li]{huang2024knowledge}
Lili Huang, Yiming Cao, Pengcheng Jia, Chenglong Li, Jin Tang, and Chuanfu Li.
\newblock Knowledge-guided cross-modal alignment and progressive fusion for chest x-ray report generation.
\newblock \emph{IEEE Transactions on Multimedia}, 2024.

\bibitem[Kang et~al.(2023)Kang, Chikontwe, Kim, Jin, Adeli, Pohl, and Park]{kang2023one}
Myeongkyun Kang, Philip Chikontwe, Soopil Kim, Kyong~Hwan Jin, Ehsan Adeli, Kilian~M Pohl, and Sang~Hyun Park.
\newblock One-shot federated learning on medical data using knowledge distillation with image synthesis and client model adaptation.
\newblock In \emph{International Conference on Medical Image Computing and Computer-Assisted Intervention}, pp.\  521--531. Springer, 2023.

\bibitem[Karimireddy et~al.(2020)Karimireddy, Kale, Mohri, Reddi, Stich, and Suresh]{karimireddy2020scaffold}
Sai~Praneeth Karimireddy, Satyen Kale, Mehryar Mohri, Sashank Reddi, Sebastian Stich, and Ananda~Theertha Suresh.
\newblock Scaffold: Stochastic controlled averaging for federated learning.
\newblock In \emph{International conference on machine learning}, pp.\  5132--5143. PMLR, 2020.

\bibitem[Ke et~al.(2021)Ke, Shen, and Lu]{ke2021style}
Jing Ke, Yiqing Shen, and Yizhou Lu.
\newblock Style normalization in histology with federated learning.
\newblock In \emph{2021 IEEE 18th International Symposium on Biomedical Imaging (ISBI)}, pp.\  953--956. IEEE, 2021.

\bibitem[Kim et~al.(2024)Kim, Kim, Kim, Kim, and Park]{kim2024fedwt}
Geonhui Kim, Jiha Kim, Yongho Kim, Hwan Kim, and Hyunhee Park.
\newblock Fedwt: Federated learning with minimum spanning tree-based weighted tree aggregation for uav networks.
\newblock \emph{ICT Express}, 2024.

\bibitem[Koga(2025)]{koga2025evaluating}
Shunsuke Koga.
\newblock Evaluating chatgpt in pathology: towards multimodal ai in medical imaging.
\newblock \emph{Journal of clinical pathology}, 78\penalty0 (1):\penalty0 70--70, 2025.

\bibitem[Li et~al.(2020)Li, Sahu, Zaheer, Sanjabi, Talwalkar, and Smith]{li2020federated}
Tian Li, Anit~Kumar Sahu, Manzil Zaheer, Maziar Sanjabi, Ameet Talwalkar, and Virginia Smith.
\newblock Federated optimization in heterogeneous networks.
\newblock \emph{Proceedings of Machine learning and systems}, 2:\penalty0 429--450, 2020.

\bibitem[Lin et~al.(2020)Lin, Kong, Stich, and Jaggi]{lin2020ensemble}
Tao Lin, Lingjing Kong, Sebastian~U Stich, and Martin Jaggi.
\newblock Ensemble distillation for robust model fusion in federated learning.
\newblock \emph{Advances in neural information processing systems}, 33:\penalty0 2351--2363, 2020.

\bibitem[Maaz et~al.(2023)Maaz, Rasheed, Khan, and Khan]{maaz2023video}
Muhammad Maaz, Hanoona Rasheed, Salman Khan, and Fahad~Shahbaz Khan.
\newblock Video-chatgpt: Towards detailed video understanding via large vision and language models.
\newblock \emph{arXiv preprint arXiv:2306.05424}, 2023.

\bibitem[McMahan et~al.(2017)McMahan, Moore, Ramage, Hampson, and y~Arcas]{mcmahan2017communication}
Brendan McMahan, Eider Moore, Daniel Ramage, Seth Hampson, and Blaise~Aguera y~Arcas.
\newblock Communication-efficient learning of deep networks from decentralized data.
\newblock In \emph{Artificial intelligence and statistics}, pp.\  1273--1282. PMLR, 2017.

\bibitem[Nakano et~al.(2021)Nakano, Hilton, Balaji, Wu, Ouyang, Kim, Hesse, Jain, Kosaraju, Saunders, et~al.]{nakano2021webgpt}
Reiichiro Nakano, Jacob Hilton, Suchir Balaji, Jeff Wu, Long Ouyang, Christina Kim, Christopher Hesse, Shantanu Jain, Vineet Kosaraju, William Saunders, et~al.
\newblock Webgpt: Browser-assisted question-answering with human feedback, 2021.
\newblock \emph{URL https://arxiv. org/abs/2112.09332}, 2021.

\bibitem[Onan(2023)]{onan2023hierarchical}
Aytu{\u{g}} Onan.
\newblock Hierarchical graph-based text classification framework with contextual node embedding and bert-based dynamic fusion.
\newblock \emph{Journal of king saud university-computer and information sciences}, 35\penalty0 (7):\penalty0 101610, 2023.

\bibitem[Ouyang et~al.(2022)Ouyang, Wu, Jiang, Almeida, Wainwright, Mishkin, Zhang, Agarwal, Slama, Ray, et~al.]{ouyang2022training}
Long Ouyang, Jeffrey Wu, Xu~Jiang, Diogo Almeida, Carroll Wainwright, Pamela Mishkin, Chong Zhang, Sandhini Agarwal, Katarina Slama, Alex Ray, et~al.
\newblock Training language models to follow instructions with human feedback.
\newblock \emph{Advances in neural information processing systems}, 35:\penalty0 27730--27744, 2022.

\bibitem[Pfitzner et~al.(2021)Pfitzner, Steckhan, and Arnrich]{pfitzner2021federated}
Bjarne Pfitzner, Nico Steckhan, and Bert Arnrich.
\newblock Federated learning in a medical context: a systematic literature review.
\newblock \emph{ACM Transactions on Internet Technology (TOIT)}, 21\penalty0 (2):\penalty0 1--31, 2021.

\bibitem[Rahman et~al.(2020)Rahman, Hasan, Lee, Zadeh, Mao, Morency, and Hoque]{rahman2020integrating}
Wasifur Rahman, Md~Kamrul Hasan, Sangwu Lee, Amir Zadeh, Chengfeng Mao, Louis-Philippe Morency, and Ehsan Hoque.
\newblock Integrating multimodal information in large pretrained transformers.
\newblock In \emph{Proceedings of the conference. Association for Computational Linguistics. Meeting}, volume 2020, pp.\  2359. NIH Public Access, 2020.

\bibitem[Rakowski et~al.(2006)Rakowski, Winterkorn, Paul, Steele, Halpern, and Thiele]{rakowski2006renal}
SK~Rakowski, EB~Winterkorn, E~Paul, DJR Steele, Elkan~F Halpern, and EA~Thiele.
\newblock Renal manifestations of tuberous sclerosis complex: incidence, prognosis, and predictive factors.
\newblock \emph{Kidney international}, 70\penalty0 (10):\penalty0 1777--1782, 2006.

\bibitem[Raza(2019)]{raza2019improving}
Khalid Raza.
\newblock Improving the prediction accuracy of heart disease with ensemble learning and majority voting rule.
\newblock In \emph{U-Healthcare Monitoring Systems}, pp.\  179--196. Elsevier, 2019.

\bibitem[Reale-Nosei et~al.(2024)Reale-Nosei, Amador-Dom{\'\i}nguez, and Serrano]{reale2024vision}
Gabriel Reale-Nosei, Elvira Amador-Dom{\'\i}nguez, and Emilio Serrano.
\newblock From vision to text: A comprehensive review of natural image captioning in medical diagnosis and radiology report generation.
\newblock \emph{Medical Image Analysis}, pp.\  103264, 2024.

\bibitem[Rsna(2019)]{rsna2019rsna}
P~Rsna.
\newblock Rsna pneumonia detection challenge, 2019.

\bibitem[Schapire(2013)]{schapire2013explaining}
Robert~E Schapire.
\newblock Explaining adaboost.
\newblock In \emph{Empirical inference: festschrift in honor of vladimir N. Vapnik}, pp.\  37--52. Springer, 2013.

\bibitem[Sheller et~al.(2020)Sheller, Edwards, Reina, Martin, Pati, Kotrotsou, Milchenko, Xu, Marcus, Colen, et~al.]{sheller2020federated}
Micah~J Sheller, Brandon Edwards, G~Anthony Reina, Jason Martin, Sarthak Pati, Aikaterini Kotrotsou, Mikhail Milchenko, Weilin Xu, Daniel Marcus, Rivka~R Colen, et~al.
\newblock Federated learning in medicine: facilitating multi-institutional collaborations without sharing patient data.
\newblock \emph{Scientific reports}, 10\penalty0 (1):\penalty0 12598, 2020.

\bibitem[Singh et~al.(2020)Singh, Sengupta, and Lakshminarayanan]{singh2020explainable}
Amitojdeep Singh, Sourya Sengupta, and Vasudevan Lakshminarayanan.
\newblock Explainable deep learning models in medical image analysis.
\newblock \emph{Journal of imaging}, 6\penalty0 (6):\penalty0 52, 2020.

\bibitem[Su et~al.(2023)Su, Li, and Xue]{su2023one}
Shangchao Su, Bin Li, and Xiangyang Xue.
\newblock One-shot federated learning without server-side training.
\newblock \emph{Neural Networks}, 164:\penalty0 203--215, 2023.

\bibitem[Sui et~al.(2020)Sui, Chen, Zhao, Jia, Xie, and Sun]{sui2020feded}
Dianbo Sui, Yubo Chen, Jun Zhao, Yantao Jia, Yuantao Xie, and Weijian Sun.
\newblock Feded: Federated learning via ensemble distillation for medical relation extraction.
\newblock In \emph{Proceedings of the 2020 conference on empirical methods in natural language processing (EMNLP)}, pp.\  2118--2128, 2020.

\bibitem[Waldock et~al.(2024)Waldock, Zhang, Guni, Nabeel, Darzi, and Ashrafian]{waldock2024accuracy}
William~J Waldock, Joe Zhang, Ahmad Guni, Ahmad Nabeel, Ara Darzi, and Hutan Ashrafian.
\newblock The accuracy and capability of artificial intelligence solutions in health care examinations and certificates: Systematic review and meta-analysis.
\newblock \emph{Journal of Medical Internet Research}, 26:\penalty0 e56532, 2024.

\bibitem[Wang et~al.(2022)Wang, Pal, Yang, Kant, Zhu, and Guo]{wang2022collaborative}
Junbo Wang, Amitangshu Pal, Qinglin Yang, Krishna Kant, Kaiming Zhu, and Song Guo.
\newblock Collaborative machine learning: Schemes, robustness, and privacy.
\newblock \emph{IEEE Transactions on Neural Networks and Learning Systems}, 2022.

\bibitem[Wang et~al.(2023)Wang, Feng, Duan, Liu, Ng, et~al.]{wang2023data}
Naibo Wang, Wenjie Feng, Moming Duan, Fusheng Liu, See-Kiong Ng, et~al.
\newblock Data-free diversity-based ensemble selection for one-shot federated learning.
\newblock \emph{Transactions on Machine Learning Research}, 2023.

\bibitem[Wang et~al.(2024{\natexlab{a}})Wang, Deng, Feng, Fan, Yin, and Ng]{wang2024one}
Naibo Wang, Yuchen Deng, Wenjie Feng, Shichen Fan, Jianwei Yin, and See-Kiong Ng.
\newblock One-shot sequential federated learning for non-iid data by enhancing local model diversity.
\newblock In \emph{Proceedings of the 32nd ACM International Conference on Multimedia}, pp.\  5201--5210, 2024{\natexlab{a}}.

\bibitem[Wang et~al.(2024{\natexlab{b}})Wang, Chen, Chen, Wu, Zhu, Zeng, Luo, Lu, Zhou, Qiao, et~al.]{wang2024visionllm}
Wenhai Wang, Zhe Chen, Xiaokang Chen, Jiannan Wu, Xizhou Zhu, Gang Zeng, Ping Luo, Tong Lu, Jie Zhou, Yu~Qiao, et~al.
\newblock Visionllm: Large language model is also an open-ended decoder for vision-centric tasks.
\newblock \emph{Advances in Neural Information Processing Systems}, 36, 2024{\natexlab{b}}.

\bibitem[Wei et~al.(2022)Wei, Wang, Schuurmans, Bosma, Xia, Chi, Le, Zhou, et~al.]{wei2022chain}
Jason Wei, Xuezhi Wang, Dale Schuurmans, Maarten Bosma, Fei Xia, Ed~Chi, Quoc~V Le, Denny Zhou, et~al.
\newblock Chain-of-thought prompting elicits reasoning in large language models.
\newblock \emph{Advances in neural information processing systems}, 35:\penalty0 24824--24837, 2022.

\bibitem[Wei et~al.(2020)Wei, Li, Ding, Ma, Yang, Farokhi, Jin, Quek, and Poor]{wei2020federated}
Kang Wei, Jun Li, Ming Ding, Chuan Ma, Howard~H Yang, Farhad Farokhi, Shi Jin, Tony~QS Quek, and H~Vincent Poor.
\newblock Federated learning with differential privacy: Algorithms and performance analysis.
\newblock \emph{IEEE transactions on information forensics and security}, 15:\penalty0 3454--3469, 2020.

\bibitem[Wolpert(1992)]{wolpert1992stacked}
David~H Wolpert.
\newblock Stacked generalization.
\newblock \emph{Neural networks}, 5\penalty0 (2):\penalty0 241--259, 1992.

\bibitem[Wu et~al.(2024{\natexlab{a}})Wu, Xie, Zhang, Phan, Chen, Chen, and Wu]{wu2024xlip}
Biao Wu, Yutong Xie, Zeyu Zhang, Minh~Hieu Phan, Qi~Chen, Ling Chen, and Qi~Wu.
\newblock Xlip: Cross-modal attention masked modelling for medical language-image pre-training.
\newblock \emph{arXiv preprint arXiv:2407.19546}, 2024{\natexlab{a}}.

\bibitem[Wu et~al.(2024{\natexlab{b}})Wu, Pei, Han, Chen, Yao, Liu, Qian, and Guo]{wu2024fedel}
Xing Wu, Jie Pei, Xian-Hua Han, Yen-Wei Chen, Junfeng Yao, Yang Liu, Quan Qian, and Yike Guo.
\newblock Fedel: Federated ensemble learning for non-iid data.
\newblock \emph{Expert Systems with Applications}, 237:\penalty0 121390, 2024{\natexlab{b}}.

\bibitem[Wu et~al.(2021)Wu, Liu, Xie, Chow, and Wei]{wu2021boosting}
Yanzhao Wu, Ling Liu, Zhongwei Xie, Ka-Ho Chow, and Wenqi Wei.
\newblock Boosting ensemble accuracy by revisiting ensemble diversity metrics.
\newblock In \emph{Proceedings of the IEEE/CVF Conference on Computer Vision and Pattern Recognition}, pp.\  16469--16477, 2021.

\bibitem[Yang et~al.(2023)Yang, Shi, Wei, Liu, Zhao, Ke, Pfister, and Ni]{yang2023medmnist}
Jiancheng Yang, Rui Shi, Donglai Wei, Zequan Liu, Lin Zhao, Bilian Ke, Hanspeter Pfister, and Bingbing Ni.
\newblock Medmnist v2-a large-scale lightweight benchmark for 2d and 3d biomedical image classification.
\newblock \emph{Scientific Data}, 10\penalty0 (1):\penalty0 41, 2023.

\bibitem[Yang et~al.(2024)Yang, Su, Li, and Xue]{yang2024exploring}
Mingzhao Yang, Shangchao Su, Bin Li, and Xiangyang Xue.
\newblock Exploring one-shot semi-supervised federated learning with pre-trained diffusion models.
\newblock In \emph{Proceedings of the AAAI Conference on Artificial Intelligence}, volume~38, pp.\  16325--16333, 2024.

\bibitem[Ye et~al.(2023)Ye, Xu, Wang, Xu, Chen, and Wang]{ye2023feddisco}
Rui Ye, Mingkai Xu, Jianyu Wang, Chenxin Xu, Siheng Chen, and Yanfeng Wang.
\newblock Feddisco: Federated learning with discrepancy-aware collaboration.
\newblock In \emph{International Conference on Machine Learning}, pp.\  39879--39902. PMLR, 2023.

\bibitem[Zeiser(2024)]{zeiser2024multsurv}
Felipe~Andr{\'e} Zeiser.
\newblock Multsurv: a multimodal deep learning model for hospitalized patients survival analysis in the contexto of a pandemic.
\newblock 2024.

\bibitem[Zhang et~al.(2022)Zhang, Chen, Li, Lyu, Wu, Ding, Shen, and Wu]{zhang2022dense}
Jie Zhang, Chen Chen, Bo~Li, Lingjuan Lyu, Shuang Wu, Shouhong Ding, Chunhua Shen, and Chao Wu.
\newblock Dense: Data-free one-shot federated learning.
\newblock \emph{Advances in Neural Information Processing Systems}, 35:\penalty0 21414--21428, 2022.

\bibitem[Zhou(2024)]{zhou2024automated}
Luping Zhou.
\newblock Automated medical report generation and visual question answering.
\newblock In \emph{Proceedings of the 1st International Workshop on Multimedia Computing for Health and Medicine}, pp.\  3--4, 2024.

\bibitem[Zhu et~al.(2023)Zhu, Chen, Shen, Li, and Elhoseiny]{zhu2023minigpt}
Deyao Zhu, Jun Chen, Xiaoqian Shen, Xiang Li, and Mohamed Elhoseiny.
\newblock Minigpt-4: Enhancing vision-language understanding with advanced large language models.
\newblock \emph{arXiv preprint arXiv:2304.10592}, 2023.

\end{thebibliography}
\bibliographystyle{tmlr}


\end{document}